\documentclass[letterpaper]{article} 
\usepackage{aaai23}  
\usepackage{times}  
\usepackage{helvet}  
\usepackage{courier}  
\usepackage[hyphens]{url}  
\usepackage{graphicx} 
\usepackage{natbib}  
\usepackage{caption} 
\usepackage{algorithm}
\usepackage{algorithmic}
\usepackage{booktabs}
\usepackage{pifont}
\usepackage{graphicx}
\usepackage{bbding}
\usepackage{subfig}
\usepackage{multirow}
\usepackage{multicol}
\usepackage{amsmath}
\usepackage{amsfonts}
\usepackage{enumerate}

\usepackage{newfloat}
\usepackage{listings}
\DeclareCaptionStyle{ruled}{labelfont=normalfont,labelsep=colon,strut=off} 
\floatstyle{ruled}
\newfloat{listing}{tb}{lst}{}
\floatname{listing}{Listing}
\title{Visual Subtitle Feature Enhanced Video Outline Generation}
\author{
    Qi Lv\textsuperscript{\rm 1}\thanks{This work was done when Qi Lv interned at Baidu.}, 
    Ziqiang Cao\textsuperscript{\rm 1}, 
    Wenrui Xie\textsuperscript{\rm 2},
    Derui Wang\textsuperscript{\rm 2},
    Jingwen Wang\textsuperscript{\rm 2},
    Zhiwei Hu\textsuperscript{\rm 2}, \\
    Tangkun Zhang\textsuperscript{\rm 2},
    Ba Yuan\textsuperscript{\rm 2},
    Yuanhang Li\textsuperscript{\rm 2},
    Min Cao\textsuperscript{\rm 1},
    Wenjie Li\textsuperscript{\rm 3},
    Sujian Li\textsuperscript{\rm 4},
    Guohong Fu\textsuperscript{\rm 1}
}
\affiliations{
    \textsuperscript{\rm 1}Institution for Artificial Intelligence, Soochow University, China\\
    \textsuperscript{\rm 2}Baidu Inc., Beijing, China\\
    \textsuperscript{\rm 3}The Hong Kong Polytechnic University\\
    \textsuperscript{\rm 4}Key Laboratory of Computational Linguistics, Peking University, MOE, China\\

    20205227047@stu.suda.edu.cn, \{zqcao, mcao\}@suda.edu.cn, ghfu@hotmail.com \\
    \{xiewenrui01, wangderui, wangjingwen03, huzhiwei03, zhangtangkun\}@baidu.com \\
    \{yuanba, liyuanhang\}@baidu.com \\
    cswjli@comp.polyu.edu.hk, lisujian@pku.edu.cn \\
}

\usepackage{bibentry}

\begin{document}

\maketitle

\begin{abstract}
With the tremendously increasing number of videos, there is a great demand for techniques that help people quickly navigate to the video segments they are interested in.
However, current works on video understanding mainly focus on video content summarization, while little effort has been made to explore the structure of a video.
Inspired by textual outline generation, we introduce a novel video understanding task, namely video outline generation (VOG).
This task is defined to contain two sub-tasks: (1) first segmenting the video according to the content structure and then (2) generating a heading for each segment.
To learn and evaluate VOG, we annotate a 10k+ dataset, called \textbf{DuVOG}.
Specifically, we use OCR tools to recognize subtitles of videos.
Then annotators are asked to divide subtitles into chapters and title each chapter.
In videos, highlighted text tends to be the headline since it is more likely to attract attention.
Therefore we propose a \textbf{V}isual \textbf{S}ubtitle feature \textbf{E}nhanced video outline generation model (\textbf{VSENet}) which takes as input the textual subtitles together with their visual font sizes and positions.
We consider the VOG task as a sequence tagging problem that extracts spans where the headings are located and then rewrites them to form the final outlines.
Furthermore, based on the similarity between video outlines and textual outlines, we use a large number of articles with chapter headings to pretrain our model.
Experiments on DuVOG show that our model largely outperforms other baseline methods, achieving 77.1 of F1-score for the video segmentation level and 85.0 of ROUGE-L$_{F0.5}$ for the headline generation level\footnote{\url{https://github.com/Aopolin-Lv/VSENet}}.
\end{abstract}

\section{Introduction}
Over the past few years, the number of videos has been growing at an incredible rate due to the progress of portable video filming devices and the prevalence of video-sharing platforms.
To help people understand videos, numerous video summarization technologies have sprung up.
For example, some works \cite{li2015summarization, song2015tvsum} compress the video content to a shot clip, and others \citet{shetty2016frame,venugopalan2016improving} generate natural language descriptions according to the video.
However, existing works focus on summarizing a video into a short synopsis, but little effort is made to analyze and explore the video content structure.
\begin{figure}[tp]%
    \centering
    \subfloat{
        \includegraphics[width=0.46\linewidth, height=5.7cm]{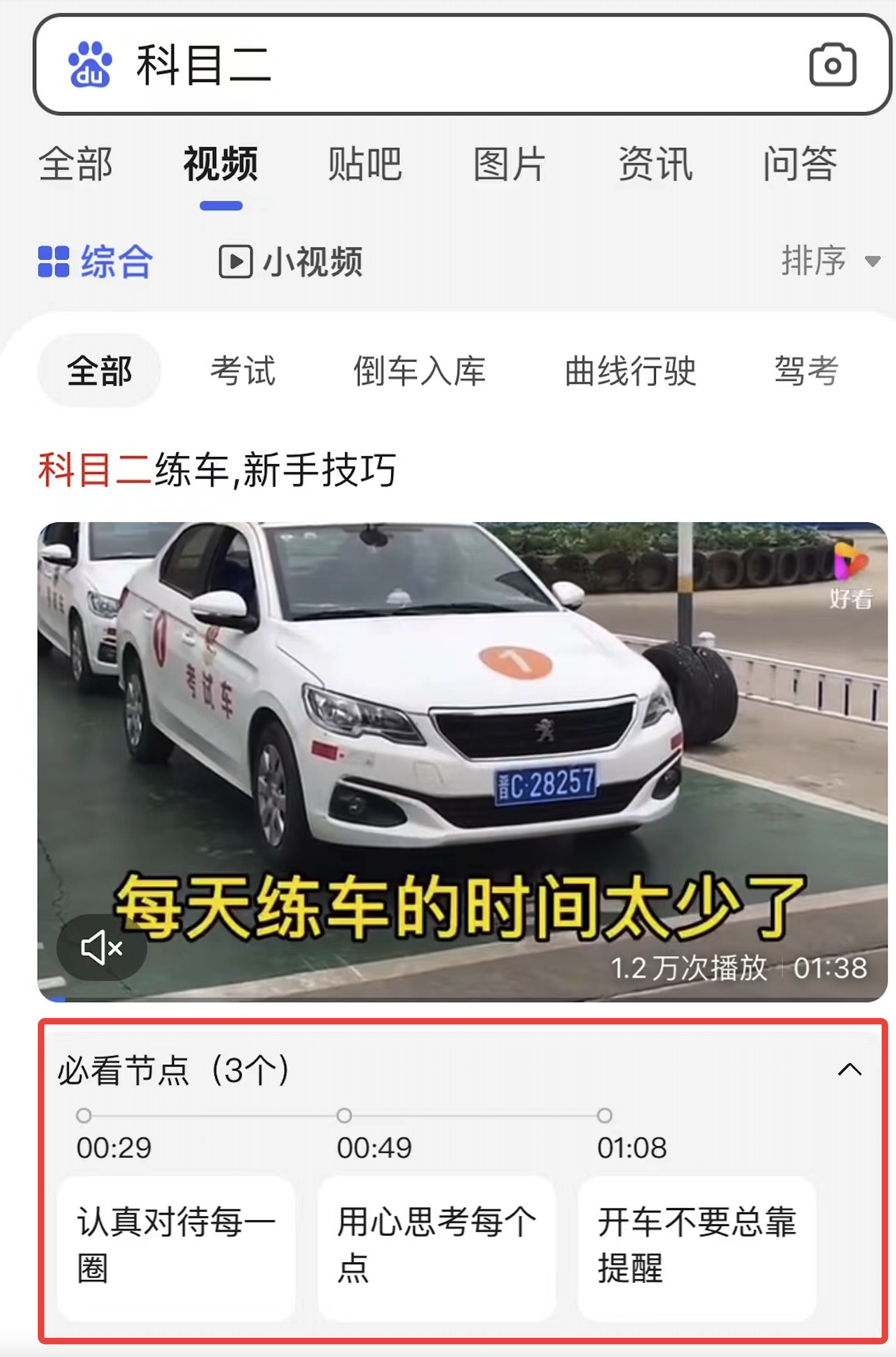}
        }
    \subfloat{
        \includegraphics[width=0.48\linewidth, height=5.9cm]{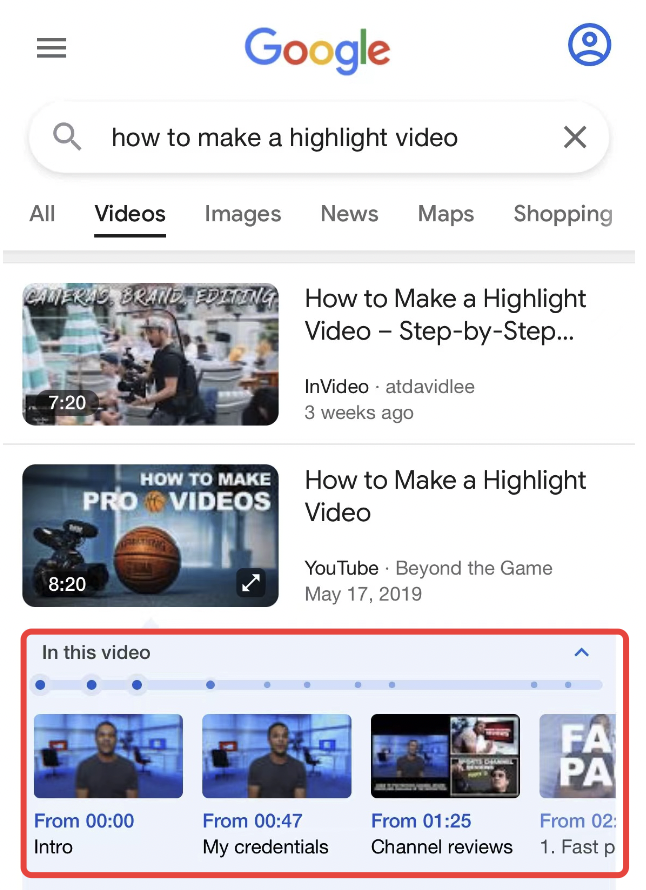}
        }
    \caption{An example of the video outline application. The left comes from Baidu search engine while the right is from that of Google.
    The red box areas belong to the video outlines.}
    \label{fig:application}
\end{figure}
Recently, many companies, such as Google and Baidu, have launched a new video feature to reveal video structures.
As shown in Figure \ref{fig:application}, they present some time stamps on the progress bar, along with the corresponding descriptions.
These time stamps and descriptions function like book chapters and chapter titles.
We call this new video feature \textbf{video outline}.
With the help of video outlines, it is more convenient for the audience to glance at the current video content and jump to wherever they are interested, thus improving the viewing experience and increasing viewer engagement.
Video outlines also facilitate platforms to understand which parts of the video are popular and raise the video quality of those parts.

However, most of the current video outlines are provided by authors artificially, which seriously hinders the promotion of this new feature.
Inspired by text outline generation~\cite{zhang2019outline} which identifies the potential sections and generates their corresponding headings, we define a novel task named video outline generation (VOG) and develop an automated solution for it.
VOG needs to segment the video content and then generate a heading for each segment.
Intuitively, this task can be regarded as a compound of video summarization~\cite{ma2002user} and video caption generation~\cite{tang2002spatial}.
However, since subtitles in video usually contain information more relevant to the structure of the video content and that image key frame extraction is time-consuming, we mainly choose to use the subtitles as the basis for VOG.

In order to develop and evaluate the VOG solutions, we annotate a new benchmark dataset called DuVOG.
Concretely, we collect 10k+ Chinese videos on generic topics from Baidu haokan videos\footnote{\url{https://haokan.baidu.com}} and adopt a public OCR tool, PaddleOCR\footnote{\url{https://github.com/PaddlePaddle/PaddleOCR}}, to recognize subtitles.
For each video, its subtitles are concatenated with the time points where they appear to form a complete text description for subsequent annotation.
We ask the annotators to understand the video content thoroughly first and then segment the video subtitles by marking the time points that denote segment boundaries.
Next, the annotators are instructed to write a concise heading for each segment.

\begin{figure}[tp]
    \centering
    \includegraphics[width=\linewidth]{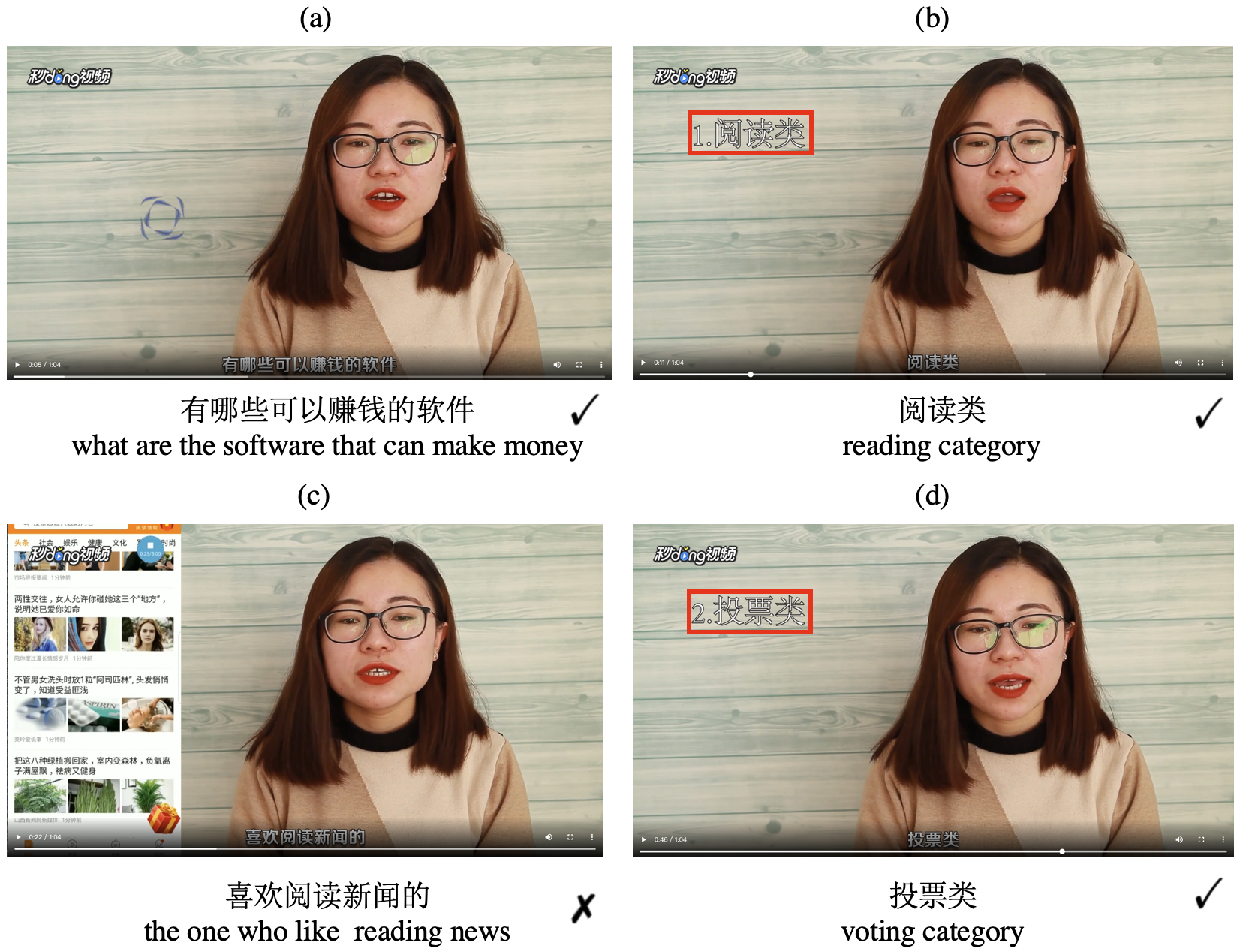}
    \caption{Pictures (a) - (d) show the frames where ``\ding{51}/ \ding{55}'' denotes the subtitle whether contains the headline.
    The red box denotes the highlighted parts.
    }
    \label{fig:case study}
\end{figure}

As Figure \ref{fig:case study} shows, to attract attention, headlines implied in video subtitles usually have distinct visual features, manifesting as their outstanding visual font size and placement.
This kind of visual features may benefit the detection of important segments.
To this end, we propose VSENet \footnote{This model has been used in an industrial system and more descriptions will be presented in the final version.}, which adopts the visual subtitle feature to enhance the model capability of video outline generation.
Our VSENet leverages the textual subtitles as well as their visual information (i.e., the visual font sizes and positions of subtitles) as input.
We formulate this VOG task as a sequence tagging problem and apply a two-stage framework to learn video segmentation and outline rewriting.
In the segmentation stage, we encode the text utilizing BERT \cite{devlin2018bert} and inject visual subtitle features to extract spans where the headings are located.
In the rewriting stage, we remove the redundant characters in the extracted spans using a Seq2Edit model LaserTagger \cite{malmi2019encode}.
In addition, based on the similarity with the text outlines, we craw a large scale articles with chapter headings and pretrain our model to detect the headings.

To summarize, our contributions are as follows:
\begin{itemize}
\item  We introduce a novel task named video outline generation for revealing the video content structure and annotate a benchmark dataset for this task.
\item We propose a visual subtitle feature enhanced model (VSENet) which adopts the textual subtitles and visual subtitle features for video outline generation.
\item We investigate an article heading detection pretraining task to improve the competence of structure understanding for our model.
\end{itemize}

\section{Related Work}
To the best of our knowledge, video outline generation is a novel task.
Therefore, in this section, we briefly introduce three related tasks: 1) video summarization \cite{ma2002user}; 2) video caption generation \cite{tang2002spatial}; 3) textual outline generation \cite{zhang2019outline}.
The former two tasks are classic video understanding tasks, while the latter one is to learn the textual structure.

\paragraph{Video Summarization}
Video summarization aims to generate a short synopsis that summarizes the video content by selecting its most informative and important shots.
Existing methods could be cast into three categories: unsupervised methods, weakly supervised methods, and supervised methods.
For the first category, some works \cite{zhou2018deep, zhang2019dtr, yaliniz2021using} attempted to reconstruct and discriminate the original video by the techniques of GAN and VAE.
For the second category, representative methods \cite{song2015tvsum, panda2017weakly} leverage some auxiliary information, including web-image priors, video titles, and video categories.
Although unsupervised and weakly supervised methods have achieved remarkable progress, they can not learn well from human annotated summaries.
For supervised learning, some deep learning methods \cite{rochan2018video, hussain2019cloud, zhu2020dsnet, narasimhan2021clip} take this a big step forward.
They usually form this task as a key frame extraction problem and employ technologies like CNN, RNN, LSTM and Transformer.

\paragraph{Video Caption Generation}
Video caption generation aims to describe the video content using natural language sentences automatically.
Early works \cite{lee2008save, thomason2014integrating} usually pre-define a set of entities and events separately, then form template sentences with the detected entities and events.
Due to the detection action for each entity, these methods own a high computational cost.
Meanwhile, they do not have ability to describe the open domain videos with unseen entities.
With the rapid development in deep learning, the neural captioning methods \cite{shetty2016frame, zhou2020unified, tang2021clip4caption, zhang2021open} that follow the encoder-decoder framework have risen to prominence.

\paragraph{Textual Outline Generation}
The goal of textual outline generation is to produce a structure output with short descriptions (i.e., headlines).
\citet{zhang2019outline} proposed a hierarchical LSTM model with an attention mechanism to improve the consistency and eliminate duplication between section headings.
With a similar modeling method, another work \cite{barrow2020joint} adopted the segment pooling LSTM model to split a document and label segments.

\section{VOG Dataset Construction}
In order to study and evaluate the VOG task, we build a new benchmark dataset named DuVOG.
We collect 10k+ raw videos with more than 500 pageviews from BAIDU haokan videos.
Most of these videos belong to the knowledge explanation/evaluation type, as their content has a clear structure.
The topics of video content are about diverse domains, including education, science, etc.

For the aim of the VOG task, we first extract images for each video at two frames per second and adopt PaddleOCR, an open source OCR tool, to recognize all text in the frames.
Then we remove the duplicate recognized sentences and stitch them with their time points together to compose the final context for follow-up annotation.
It is worth noting that although we have proofread the recognition result of each frame against its adjacent frames, some errors such as missing characters, duplicate characters and spelling errors will inevitably occur during the OCR recognition process.

The following sections present the details about the annotation target, criteria and procedure.
Finally, we describe the characteristics of this dataset.

\subsubsection{Annotation Procedure}
In order to ensure the annotation quality, we recommend that the annotators first understand the video content thoroughly, and then divide it into segments.
For each segment, annotators need to mark the segmentation points and their outlines reference.
Meanwhile, they are required to write the final outlines according to the reference.
We recruited four general annotators and two high-rank reviewers in all.
The high-rank reviewers are required to proofread the annotation results of the general annotators.
They all undergo professional labeling training and assessment.
The annotation accuracy rates of these higher-rank reviewers and general annotators are 95\%+ and 85\%+, respectively.
The whole annotation process lasted for one month and we checked the annotation results once per week.
Furthermore, to make the annotation results more in line with our expectations, when it comes to some complex or uncertain samples, we collect them and discuss regularly to improve the annotation criterion.

\begin{figure}[htbp]
    \centering
    \small
    \includegraphics[width=\linewidth]{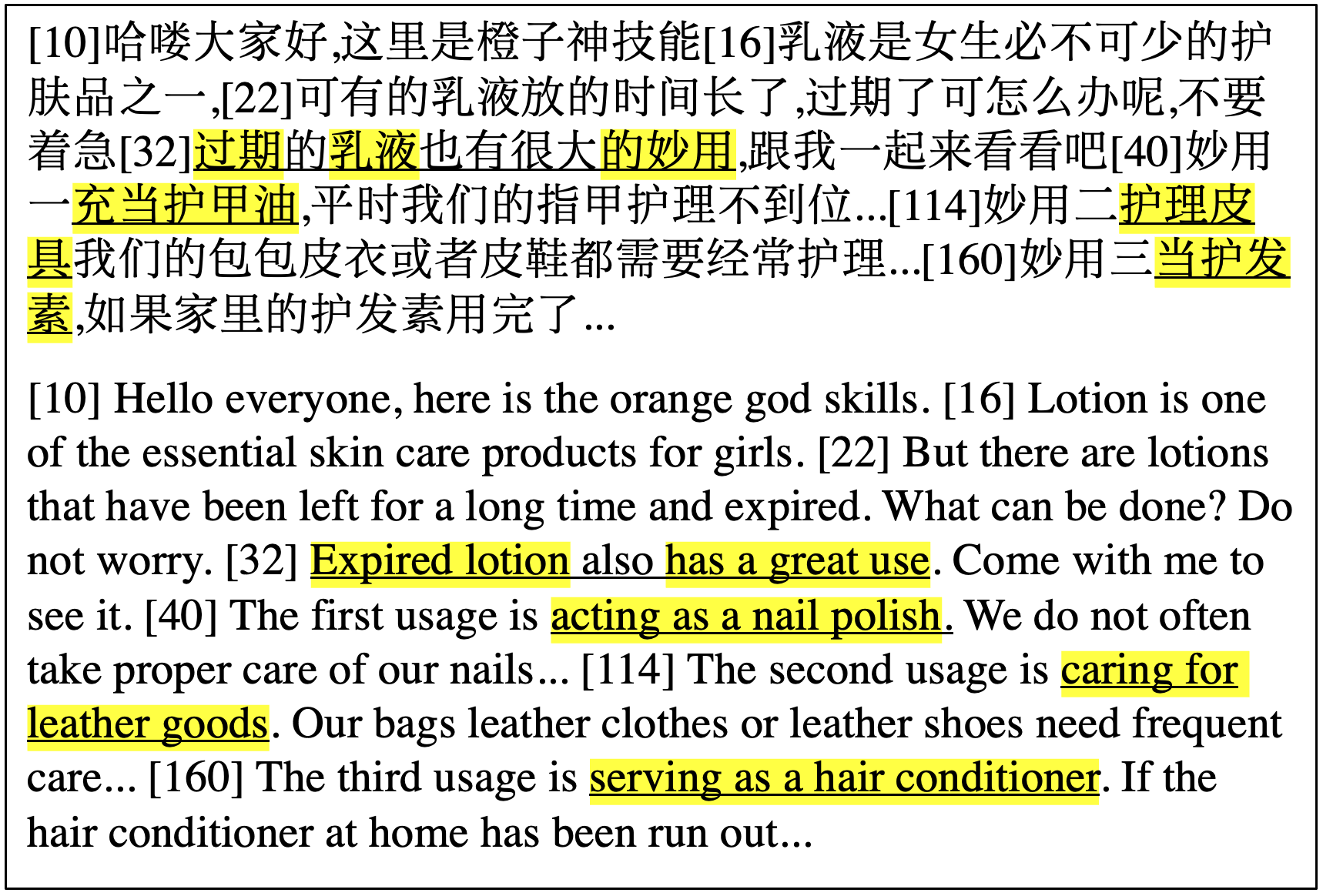}
    \caption{An example in our DuVOG dataset.
    The marker ``[x]'' in front of the sentence indicates the timestamp when the sentence appears.
    The underlined sentences represent outline references, while the highlighted parts are the final outline results.}
    \label{fig:dataset_sample}
\end{figure}
\subsubsection{Annotation Criteria}
We provide annotators with the original videos and their subtitle sequences.
The annotation criteria for segmentation points and candidate outlines are presented in details as follows:
\begin{itemize}
    \item Segmentation Point:
    Annotators are required to split the entire video subtitle sequence to segments according to its content and then select the corresponding time point as the segmentation point.
    It is only annotated at a time point that appears at the beginning of a segment.
    When it comes to more than one outline in a segment, we only record the time point once.
    \item Outline:
    It is comprised of two periods: reference span annotation and outline rewriting.
    In the former part, we need to locate the shortest span of each segment which maintain its originally complete semantics.
    The extracted spans must be continuous in the original text.
    In the latter part, the annotators are required to rewrite the outlines according to the reference, such as removing the redundant signals and characters, and adjusting the order of characters.
    For brevity purposes, they are not allowed to add extra characters when rewriting.
    Figure \ref{fig:dataset_sample} shows an example.
    Outline references are underlined, while their rewriting results are highlighted.
    For the first outline reference among them, the annotators remove the redundant characters ``also'' for simplicity.
\end{itemize}

\begin{table}[htbp]
  \centering
  \small
    \begin{tabular}{|l|r|r|r|}
    \hline
          & \multicolumn{1}{r|}{Train} & \multicolumn{1}{r|}{Dev} & \multicolumn{1}{r|}{Test} \\
    \hline
    Size  & 10,585 & 158   & 160 \\
    \hline
    Median video time (s) & 130.0 & 94.5 & 84.5 \\
    \hline
    Median text length & 654.0 & 421.0 & 384.0 \\
    \hline
    \# Avg. outlines & 6.2   & 5.6   & 5.7 \\
    \hline
    Avg. outline length & 6.9   & 5.9   & 5.4 \\
    \hline
    Rewriting ratio &  21.2\%  & 30.8\%  & 28.0\% \\
    \hline
    Remove ratio &  99.6\%  & 99.6\%  & 99.2\% \\
    \hline
\end{tabular}%
  \caption{Statistics of the DuVOG dataset. ``s'' stands for time in seconds.}
  \label{tab:dataset statistics}%
\end{table}%
\subsection{Dataset Characteristics}
The videos comprised in this dataset have two features:
a). the explicit theme.
We choose videos whose topics are most relevant to the introduction, evaluation or interview, so the video content is mainly based on a distinctive theme.
In this video, the content is usually elaborated around a particular thing/event from the first or third person perspective.
Therefore these video expressions often contain duplicated oral terms or non-essential conjunctions, which disrupt the original semantics of sentences and bring a big challenge to machine comprehension;
b). the well-organized content structure.
To ensure video quality, some authors may conceive the video content and form a specific shooting script in advance.
In this way, they are able to present their ideas in a clear logic, preventing audiences from being confused.
For example, in an evaluation of the iPhone 13, the author will introduce some aspects following a relatively complete line of logic, such as appearance, performance and price.

Each video sample in our DuVOG dataset has its textual subtitles, target outlines, title and download link.
The target outline is comprised of the beginning time point (i.e., segmentation point) and its reference, where the latter can be extracted from the original text.
Table \ref{tab:dataset statistics} shows the overall statistics of our DuVOG dataset.
As can be seen, the videos are not long in duration while there are 6 outlines per video on average.
Meanwhile, the outlines are concise, with an average length of 6.1.
The development set and test set are not randomly divided. 
Instead, we select some specific videos from different topics.
Therefore, their average length differs from that of the training set.
It is worth noting that the percentage of rewriting is not high, which indicates that its difficulties are mainly focused on segmentation.
Among the rewriting spans, the remove operation accounts for the most percentage.
See more details about DuVOG in Appendix.

\section{Visual Subtitle Feature Enhanced VOG Model}
\subsection{Task Definition}
The VOG task aims at revealing the inherent content structure of videos.
Formally, given a video $v$, this task expects to divide the video content into different segments $(t_1, t_2, \ldots, t_m)$, and generate their corresponding video outlines $(o_1, o_2, \ldots, o_m)$ simultaneously,
\begin{align}
    g(v) = ((t_1, o_1), (t_2, o_2), \ldots, (t_m, o_m)),
\end{align}
where $m$ denotes the number of outlines.

\subsection{Model Overview}
We leverage the textual subtitle of videos to represent input video, along with the visual subtitle information including the font sizes and positions of subtitles.
Formally, we regard the VOG task as a sequence tagging problem and we intuitively divide it into two stages: the span extraction and the outline rewriting.
The span refers to the continuous context which contains the headline.
With a generated outline, we trace back to the time point when it appears, and then consider the time point as the segmentation point.
Figure \ref{fig:model} shows our model's overview.
We also attempt the one stage training strategy which predicts the non-continuous tags.
More about this experiment is presented in Section \ref{experiment}.
In the following parts, we describe these two stages in detail.

\subsection{Span Extraction}
The goal of this stage is to detect and extract the important spans where the headings are located.
We consider this period as a sequence tagging task where the input is the sequence of subtitle characters and the output is the sequence of BIO (Beginning, Inside, Outside) tags.
\subsubsection{Input}
We remove the timestamp characters from the subtitles for the purpose of preserving the complete contextual semantics of the sentence.
Each subtitle is then combined with a comma.
As Figure \ref{fig:input_sample} shows, ``[32]'' is removed while these two adjacent subtitles are combined with a comma.

\subsubsection{Output}
\begin{figure}[ht]
    \small
    \centering
    \includegraphics[width=\linewidth]{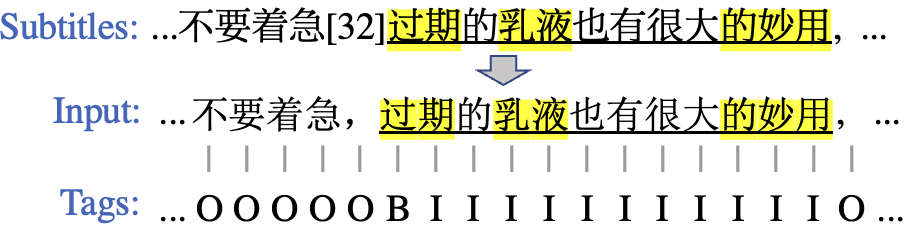}
    \caption{An example of conversion from a subtitle to the input data, along with its corresponding tags.}
    \label{fig:input_sample}
\end{figure}

According to the start/end index of outline reference in the dataset, we automatically generate the continuous token tags from \{B,I,O\} which used in the named entity recognition (NER) task \cite{zeng2014relation}.
The tag of token in outline reference is set to ``B/I'' and others are formed as ``O''.
In Figure \ref{fig:input_sample}, it can be seen that the target we focus on in this period is the underlined span, regardless of highlighted characters.

\subsubsection{Model Description}\label{model_description}
\begin{figure*}[tp]
    \centering
    \resizebox{\linewidth}{!}{
    \includegraphics{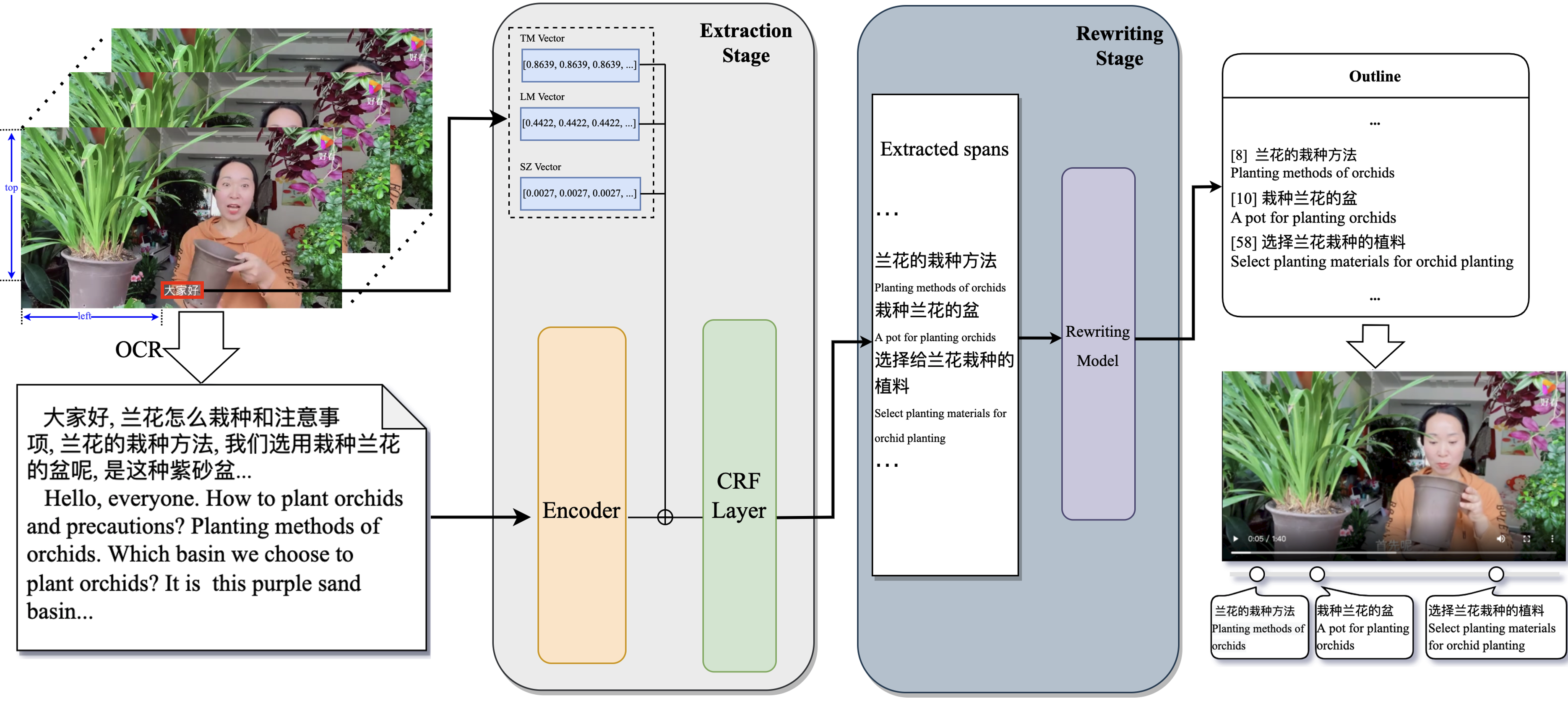}}
    \caption{Our framework overview. 
    In the top left part, the red area represents the text box while the top margin, left margin of the box are in blue.}
    \label{fig:model}
\end{figure*}

Depending on when each subtitle appears, we extract its corresponding frame.
As is shown in the top left subfigure of Figure \ref{fig:model}, we earn the position of the subtitle box, as well as its size.
Based on the size of the video image and the information of the text box, we can calculate the relative position of the text box and the area of its individual character.

Formally, we first obtain the height $D_h$ and width $D_w$ of the original video image.
Then, according to the timestamps of subtitles, we extract their corresponding frames and recognize the subtitle boxes $\mathbf{B}=\{b_1, b_2, \ldots, b_m\}$ where a box $b_i$ is comprised of tokens $\{x_p, x_{p+1}, \ldots, x_{p+l}\}$.
We consider the box as a whole, then the relative position (i.e., $v^{TM}_k$ and $v^{LM}_k$) and font size $v^{SZ}_k$ of each token in the box is:
\begin{align}
    v^{TM}_k &= \frac{d^{TM}_i}{D_h}, v^{LM}_k = \frac{d^{LM}_i}{D_w}, \\
    v^{SZ}_k &= \frac{d^h_i \cdot d^w_i}{D_h \cdot D_w \cdot l},
\end{align}
where $d^{TM}_i$, $d^{LM}_i$, $d^h_i$ and $d^w_i$ denote the top margin, left margin, height and width of box $b_i$;
$k\in[p, p+l]$.

Following most NLP approaches \cite{xu2019bert}, we adopt the pretrained language model BERT as our text encoder and truncate the part that exceeds the length.
Note, we have also attempted XLNet \cite{yang2019xlnet} with no input length limit as the text encoder (see details in Section \ref{experiment}).
Then we combine the textual embedding with the visual representation by a gate mechanism.
To learn the dependency between labels, we add the conditional random fields (CRF) layer \cite{lafferty2001conditional} following \citet{huang2015bidirectional}.
We obtain the prediction $\mathbf{\hat{Y}}$ as follows:
\begin{align}
    \mathbf{H}^t &= \operatorname{BERTEncdoer}(\mathbf{X}),\\
    \mathbf{H}^v &= [\mathbf{V}^{TM}, \mathbf{V}^{LM}, \mathbf{V}^{SZ}], \\
    \mathbf{H}^g &= \sigma{(\mathbf{W}\mathbf{H}^v + b)} \odot \mathbf{H}^t, \\
    \mathbf{\hat{Y}} &= \operatorname{CRFLayer}(\mathbf{H}^g),
\end{align}
where $\mathbf{W}$ and $b$ is learnable parameters.
$\sigma$ is a nonlinear activation function, which is a ReLU function in our implementation.
``$\odot$'' represents element-wise multiplication.

\subsubsection{Heading Detection Pretraining}
Due to the limited amount of data for the VOG task, we use articles with outlines to additionally pretrain the base model BERT based to improve its ability for chapter segmentation.

\begin{figure}[tp]
    \centering
    \includegraphics[width=0.8\linewidth]{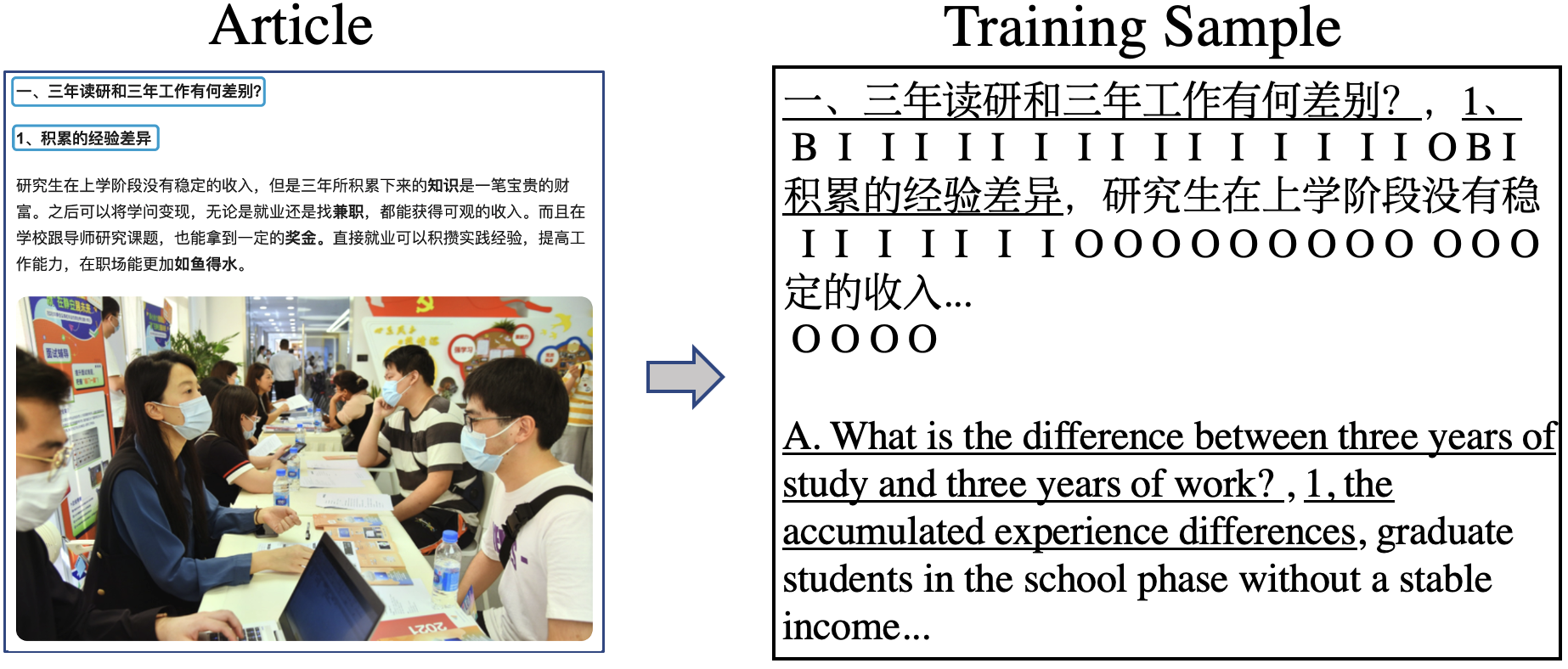}
    \caption{An example of the conversion from article to training samples.
    Sentences surrounding by the blue box in article (left) represent the headlines, corresponding to the underlined text in the training sample (right).}
    \label{fig:pre_input}
\end{figure}

Specifically, we collect articles from Baidu BAIJIAHAO\footnote{\url{https://baijiahao.baidu.com}}, and obtain the headlines from the HTML tag information.
Next, we discard the articles with less than 3 headings to ensure the structure completeness.
For the rest approximately 3.3 million articles, we concatenate each main context and headings with commas to form a long sequence.
We mapped the formulated text sequences to labels for follow-up self-supervised learning, depending on whether the character belonged to a headline.
As Figure \ref{fig:pre_input} shows, if the character is part of a heading, its corresponding tag will be set to ``B/I'', otherwise ``O'', like the tags in the VOG task.

Similar to our span extraction period, it aims to detect heading segments in context.
Considering that the model can learn from a large scalar training data and the CRF layer may cause additional overhead, we formulate this task as a token classification task with three labels \{B, I, O\}.
Formally, given a text sequence $\mathbf{X}=\{x_1, x_2, \ldots, x_n\}$, we obtain the prediction $\mathbf{\hat{Y}}=\{\hat{y_1}, \hat{y_2}, \ldots, \hat{y_n}\}$ as followed:
\begin{align}
\small
    \mathbf{H}^t &= \operatorname{BERTEncdoer}(\mathbf{X}),\\
    \mathbf{P} &= \operatorname{Softmax}(\mathbf{H}^t), \\
    \mathbf{\hat{Y}} &= \operatorname{argmax}(\mathbf{P}).
\end{align}

\subsection{Outline Rewriting}
The extracted spans are available as a summary in many cases.
To generate a concise outline, we feed the extracted spans to a rewriting model to delete the redundant characters.
Motivated by the similar purpose of grammatical error correction \cite{dahlmeier2012better}, we apply a simple yet effective Seq2Edit model, LaserTagger \cite{malmi2019encode}, to rewrite the extracted spans.
Since character deletion accounts for the majority of rewriting, we only retain the ``Keep'' and ``Delete'' operations as the prediction tags.
Notably, there are two decoding methods in LaserTagger, and we leverage the autoregressive Transformer decoder.

Overall, we extract segments with headlines and view timestamps they appeared as segmentation points of the video structure.
Then the extracted spans are rewritten to obtain final outlines.
\section{Experiment}\label{experiment}
\subsection{Experimental Settings}
To evaluate the performance of our model, we conduct experiments on our DuVOG benchmark dataset.
We implement our model based on huggingface's pytorch implementation of transformers\footnote{\url{https://github.com/huggingface/transformers}} and google's tensorflow implementation of lasertagger\footnote{\url{https://github.com/google-research/lasertagger}}.
The pre-trained bert-base-chinese is employed to initialize the BERT model.
For the extraction stage, we set the training batch size to 32 and the AdamW \cite{loshchilov2017decoupled}  optimizer is applied.
The initial learning rate of model parameters is set to 3e-5, except for that of the CRF layer which is set to 1e-2.
We set the training epoch of this stage to 20 empirically.
Concerning the heading detection pretraining, the training batch size is 1280 and the training steps are 50k including 2k warmup steps.
The learning rate of AdamW algorithm is set as 5e-5.
For the rewriting stage, we set the training batch size and the optimizer the same as that in extraction stage.
The training epoch is set to 10 empirically.
The experiment is conducted on a Tesla A100 GPU card with a memory size of 40G.

\subsection{Evaluation Metrics}
To measure the quality of generated video outlines, we introduce the segmentation level metric and headline generation level metric.
In terms of segmentation, we first count the segmentation points and then calculate their precision, recall and F1 score.
As regards to headline generation level, we adopt the automatic metric ROUGE-L for evaluation.

Similar to the grammatical error correction task \cite{zhang2022mucgec}, precision is more important than recall for rewriting.
Thus we adopt ROUGE-L$_{F0.5}$ as the main metric for generation evaluation.
Notably, the outline will be meaningless once its corresponding segmentation point is wrong.
Therefore we only calculate prediction results of headings with correct segmentation points.
Finally, we conclude the model overall performance by multiplying the F1 score at segmentation level and ROUGE-L$_{F0.5}$ at headline generation level.
In addition, we perform human evaluation to ensure the prediction quality.

\subsection{Baseline Methods}
We compare four different methods:
1) \textbf{JointBC}:
This approach considers the non-continuous token tags of \{B, I, O\} as its prediction labels.
Likewise, it adopts the BERTEncoder to encode sentences and add the CRF layer to output the result. 
2) \textbf{BC-base}:
This method applies BERT as the text encoder and combines the CRF layer additionally.
3) \textbf{BC-PT}:
This model enhances its structural understanding by heading detection pretraining based on BC-base.
4) \textbf{VSEBert}:
This method utilizes BERT as the text encoder and a fully connected network as the final layer to predict results.
5) \textbf{VSENet}:
This is our proposed model.
As illustrated in Table \ref{tab:baselines}, the heading detection pretraining (PT) is adopted to enhance the model performance of BC-PT, VSEBert and VSENet.
For the latter two methods, the visual subtitle feature (VSE) is also used to improve the structure understanding capability.
As mentioned in Section \ref{model_description}, we have also tried XLNet \cite{yang2019xlnet} as the text encoder.
However, compared with above the baseline methods, its training speed is 8$\times$ slower.
Meanwhile, we have attempted to use DSNet \cite{zhu2020dsnet} to extract key frames in videos.
But it also costs much time to extract visual features

To show the difference between joint and pipeline approaches, we compare JointBC and BC-base.
For the aim of verifying the effectiveness of PT, we compare BC-base and BC-PT.
We made a comparison between VSEBert and VSENet to reveal the necessity of the CRF layer.

In addition, we use different settings in the rewriting stage: 
1) \textbf{w/o rewriting}:
It means that the extraction results are directly used to be the final outlines;
2) \textbf{w/ BC$_{r}$}:
It forms the B/I/O as its tags, where ``B/I'' corresponds to the redundant span and ``O'' corresponds to the in-hold character.
Meanwhile, it adopts a BERTEncoder combined with the CRF layer as the base model;
3) \textbf{w/ LT}:
This is our approach to rewrite the span by using LaserTagger.
\begin{table}[hb]
\small
  \centering
    \begin{tabular}{|l|c|c|c|}
    \hline
    Baselines &  CRF & VSE   & PT \\
    \hline
    JointBC & \ding{55} &   \ding{55}   & \ding{55} \\
    BC-base & \ding{51} & \ding{55}     & \ding{55}\ \\
    BC-PT & \ding{51} & \ding{55} & \ding{51} \\
    VSEBert  & \ding{55}  & \ding{51}   & \ding{51} \\
    VSENet & \ding{51} & \ding{51}     & \ding{51} \\
    \hline
    \end{tabular}%
  \caption{Baseline methods.
  VSE/PT denotes the visual subtitle feature enhancement/heading detection pretraining.}
  \label{tab:baselines}%
\end{table}%

\subsection{Results Analysis}
\begin{table*}[ht]
\small
  \centering
    \begin{tabular}{|l|ccc|ccc|c|}
    \hline
          & \multicolumn{3}{c|}{Segmentation} & \multicolumn{3}{c|}{Generation} & Overall \\
\cline{2-8}& Pre.  & Rec.  & F1    & ROUGE-L$_{P}$  & ROUGE-L$_{R}$  & ROUGE-L$_{F0.5}$  & Score \\
    \hline
    JointBC & 73.6  & 69.7  & 71.6  & 81.2  & 93.8  & 81.6  & 58.4  \\
    \hline
    \multicolumn{1}{|l|}{BC-base} &       &       & \multicolumn{1}{r}{} &       &       & \multicolumn{1}{r}{} &  \\
    \hline
    \ \ \ w/o rewriting & \multirow{3}[2]{*}{79.0 } & \multirow{3}[2]{*}{68.9 } & \multirow{3}[2]{*}{73.6 } &  83.5     &  94.0     &  83.6     & 61.5  \\
    \ \ \ w/ BC$_{r}$ &       &       &       &   83.8    &  93.8     &  83.9     & 61.8 \\
    \ \ \ w/ LT &       &       &       &    84.1   &   93.0    &   84.1    & 61.9 \\
    \hline
    \multicolumn{1}{|l|}{BC-PT} &       &       & \multicolumn{1}{r}{} &       &       & \multicolumn{1}{r}{} &  \\
    \hline
    \ \ \ w/o rewriting & \multirow{3}[2]{*}{76.1} & \multirow{3}[2]{*}{73.9} & \multirow{3}[2]{*}{75.0} &  83.0     &  93.4     &  83.2     & 62.4  \\
    \ \ \ w/ BC$_{r}$ &       &       &       &   83.4    &   93.4    &   83.5    & 62.6 \\
    \ \ \ w/ LT &       &       &       &    83.7   &   92.3    &   83.8    & 62.8 \\
    \hline
    \multicolumn{8}{|l|}{VSEBert} \\
    \hline
    \ \ \ w/o rewriting & \multirow{3}[2]{*}{77.9} & \multirow{3}[2]{*}{70.8} & \multirow{3}[2]{*}{74.3} &  80.5 & 92.7   & 80.7  & 59.9 \\
    \ \ \ w/ BC$_{r}$ &     &  &   & 80.8  & 92.6  & 80.9  & 60.1  \\
    \ \ \ w/ LT &       &       &       &    81.1   &   91.6    &   81.1    & 60.3 \\
    \hline
    \multicolumn{8}{|l|}{VSENet} \\
    \hline
    \ \ \ w/o rewriting & \multirow{3}[2]{*}{\textbf{83.1 }} & \multirow{3}[2]{*}{\textbf{71.9 }} & \multirow{3}[2]{*}{\textbf{77.1 }} & 84.0  & \textbf{94.7 } & 84.6  & 65.2  \\
    \ \ \ w/ BC$_{r}$ &       &       &       & 84.3  & 94.6  & 84.9  & 65.5  \\
    \ \ \ w/ LT &       &       &       & \textbf{84.7 } & 93.5  & \textbf{85.0 } & \textbf{65.6}  \\
    \hline
    \end{tabular}
  \caption{Performance on the DuVOG dataset. Best results are in \textbf{bold}.
  The results of model in segmentation period are the same because the rewriting module only works in the generation period.
  Besides, there is no need to add additional rewriting periods to JointBC, since the non-contiguous tags can implement the removing operation.}
  \label{tab:main_result}%
\end{table*}%
\subsubsection{Main Result}
The overall performance of all baselines is shown in Table \ref{tab:main_result}.
As can be seen, our model obviously surpasses other methods in the overall score metric.
JointBC is even inferior to the simple model BC-base by 3 points.
The main resistance we deduced is that the segmentation requires global attention for a long sequence, while generation focuses on local attention for a short span.
We thus mainly concentrate on the two stage pipeline method.
At the segmentation level, by comparing the results of BC-base and BC-PT, we observe that our proposed heading detection pretraining has a promising advantage, achieving 1.4 points improvement.
With the help of visual caption feature enhancement, VSENet scores approximately 2 points higher than BC-PT, which verified its effectiveness.
Meanwhile, the increase over VSEBert demonstrates that the CRF layer is an essential part of our VSENet for the continuous span target.

Generally speaking, the rewriting module raises the ROUGE-L$_{F_{0.5}}$ metric result.
The human evaluation has also found that compressive rewriting improves audience experience.
Moreover, the LaserTagger shows a marginally increasing compared with BC$_{r}$.
We believe the autoregressive generation could benefit the final rewriting result.
By comparison, it is not difficult to find that the enhancement of the extraction stage would also help the results of the generation phase to some extent.

\subsubsection{Ablation Study}
To investigate the effectiveness of each component in our VSENet, we conduct several ablation experiments with the following settings:
1) \textbf{w/o VSE}: removing the visual subtitle feature;
2) \textbf{w/o PT}: removing the heading detection pretraining;
3) \textbf{w/ cat}: concatenating the textual embedding and visual subtitle feature.
As illustrated in Table \ref{tab:ablation}, we can observe that by removing the visual subtitle enhance feature, the precision drops significantly although the recall rises at the segmentation level.
The reason is that adding visual features would take an opposite effect since some videos do not have highlight parts.
In addition, heading detection pretraining plays an important role in our model due to its enhancement of structure understanding.
Another fusion method is still weaker than the gate mechanism although it achieves competitive performance.
\begin{table}[htbp]
  \centering
  \small
    \begin{tabular}{|l|ccc|c|}
    \hline
          & \multicolumn{3}{c|}{Segmentation} & Overall \\
    \cline{2-5}      & Pre.  & Rec.  & F1    & Score \\
    \hline
    VSENet & \textbf{83.1}  & 71.9  & \textbf{77.1}  & \textbf{65.6}  \\
    \hline
    w/o VSE & 76.1  & \textbf{73.9}  & 75.0  & 62.8  \\
    w/o PT & 81.4  & 66.7  & 73.3  & 63.3  \\
    w/\ \ \  cat & 81.6  & 72.7  & 76.9  & 65.1  \\
    \hline
    \end{tabular}%
  \caption{Ablation experiment results.
  VSE/PT denotes the visual subtitle feature enhanced method/heading detection pretraining method.
  }
  \label{tab:ablation}%
\end{table}%

\subsubsection{Quality Evaluation}
We perform human evaluation to ensure that our increase in overall scores is also followed by an increase in  machine understanding of video content structure.
We show the original videos, their subtitles, the ground truth in test set, as well as outlines generated by BC-Base and our VSENet side by side to a human evaluator.
Two scores from 0 to 3 are then assigned to each video outline, one for segmentation (how accurate the segmentation boundary position) and one for generation (how quality of the headline is).
Each summary is rated by 4 different human evaluators and the results are averaged across all examples and evaluators.
The result in Table \ref{tab:human evaluation} demonstrates that VSENet has a better understanding of the video content architecture which is consistent with the higher overall score.

\begin{table}[htbb]
  \centering
  \small
    \begin{tabular}{|l|c|c|}
    \hline
          & \multicolumn{1}{l|}{Segmentation} & \multicolumn{1}{l|}{Generation} \\
    \hline
    BC-base    &  1.54    & 1.43 \\
    \hline
    VSENet &   1.68    & 1.61 \\
    \hline
    \end{tabular}%
  \caption{The result of human evaluation.}
  \label{tab:human evaluation}%
\end{table}%


\section{Conclusion}
In this paper, we introduce a novel and practical video understanding task named video outline generation (VOG).
For evaluating the VOG task, we annotate a 10k+ Chinese video dataset called DuVOG.
What's more, we propose a visual subtitle feature enhanced model (VSENet) which adopts the textual subtitles and visual subtitle features simultaneously to segment the video content and generate a headline for each segment.
To enhance structure understanding, we pretrain our base model BERT with a large scalar articles with headings.
Experimental results reveal the effect of our VSENet.
Since this task is first introduced, there are many potential treasures to investigate, such as the extraction and fusion method of multi-modality information in videos.

\bibliography{aaai23.bib}

\begin{thebibliography}{31}
\providecommand{\natexlab}[1]{#1}

\bibitem[{Barrow et~al.(2020)Barrow, Jain, Morariu, Manjunatha, Oard, and
  Resnik}]{barrow2020joint}
Barrow, J.; Jain, R.; Morariu, V.; Manjunatha, V.; Oard, D.~W.; and Resnik, P.
  2020.
\newblock A joint model for document segmentation and segment labeling.
\newblock In \emph{Proceedings of the 58th Annual Meeting of the Association
  for Computational Linguistics}, 313--322.

\bibitem[{Dahlmeier and Ng(2012)}]{dahlmeier2012better}
Dahlmeier, D.; and Ng, H.~T. 2012.
\newblock Better evaluation for grammatical error correction.
\newblock In \emph{Proceedings of the 2012 Conference of the North American
  Chapter of the Association for Computational Linguistics: Human Language
  Technologies}, 568--572.

\bibitem[{Devlin et~al.(2018)Devlin, Chang, Lee, and
  Toutanova}]{devlin2018bert}
Devlin, J.; Chang, M.-W.; Lee, K.; and Toutanova, K. 2018.
\newblock Bert: Pre-training of deep bidirectional transformers for language
  understanding.
\newblock \emph{arXiv preprint arXiv:1810.04805}.

\bibitem[{Huang, Xu, and Yu(2015)}]{huang2015bidirectional}
Huang, Z.; Xu, W.; and Yu, K. 2015.
\newblock Bidirectional LSTM-CRF models for sequence tagging.
\newblock \emph{arXiv preprint arXiv:1508.01991}.

\bibitem[{Hussain et~al.(2019)Hussain, Muhammad, Ullah, Cao, Baik, and
  de~Albuquerque}]{hussain2019cloud}
Hussain, T.; Muhammad, K.; Ullah, A.; Cao, Z.; Baik, S.~W.; and de~Albuquerque,
  V. H.~C. 2019.
\newblock Cloud-assisted multiview video summarization using CNN and
  bidirectional LSTM.
\newblock \emph{IEEE Transactions on Industrial Informatics}, 16(1): 77--86.

\bibitem[{Lafferty, McCallum, and Pereira(2001)}]{lafferty2001conditional}
Lafferty, J.; McCallum, A.; and Pereira, F.~C. 2001.
\newblock Conditional random fields: Probabilistic models for segmenting and
  labeling sequence data.

\bibitem[{Lee et~al.(2008)Lee, Hakeem, Haering, and Zhu}]{lee2008save}
Lee, M.~W.; Hakeem, A.; Haering, N.; and Zhu, S.-C. 2008.
\newblock Save: A framework for semantic annotation of visual events.
\newblock In \emph{2008 IEEE Computer Society Conference on Computer Vision and
  Pattern Recognition Workshops}, 1--8. IEEE.

\bibitem[{Li, Ma, and Han(2015)}]{li2015summarization}
Li, G.; Ma, S.; and Han, Y. 2015.
\newblock Summarization-based video caption via deep neural networks.
\newblock In \emph{Proceedings of the 23rd ACM international conference on
  Multimedia}, 1191--1194.

\bibitem[{Loshchilov and Hutter(2017)}]{loshchilov2017decoupled}
Loshchilov, I.; and Hutter, F. 2017.
\newblock Decoupled weight decay regularization.
\newblock \emph{arXiv preprint arXiv:1711.05101}.

\bibitem[{Ma et~al.(2002)Ma, Lu, Zhang, and Li}]{ma2002user}
Ma, Y.-F.; Lu, L.; Zhang, H.-J.; and Li, M. 2002.
\newblock A user attention model for video summarization.
\newblock In \emph{Proceedings of the tenth ACM international conference on
  Multimedia}, 533--542.

\bibitem[{Malmi et~al.(2019)Malmi, Krause, Rothe, Mirylenka, and
  Severyn}]{malmi2019encode}
Malmi, E.; Krause, S.; Rothe, S.; Mirylenka, D.; and Severyn, A. 2019.
\newblock Encode, tag, realize: High-precision text editing.
\newblock \emph{arXiv preprint arXiv:1909.01187}.

\bibitem[{Narasimhan, Rohrbach, and Darrell(2021)}]{narasimhan2021clip}
Narasimhan, M.; Rohrbach, A.; and Darrell, T. 2021.
\newblock CLIP-It! language-guided video summarization.
\newblock \emph{Advances in Neural Information Processing Systems}, 34:
  13988--14000.

\bibitem[{Panda et~al.(2017)Panda, Das, Wu, Ernst, and
  Roy-Chowdhury}]{panda2017weakly}
Panda, R.; Das, A.; Wu, Z.; Ernst, J.; and Roy-Chowdhury, A.~K. 2017.
\newblock Weakly supervised summarization of web videos.
\newblock In \emph{Proceedings of the IEEE International Conference on Computer
  Vision}, 3657--3666.

\bibitem[{Rochan, Ye, and Wang(2018)}]{rochan2018video}
Rochan, M.; Ye, L.; and Wang, Y. 2018.
\newblock Video summarization using fully convolutional sequence networks.
\newblock In \emph{Proceedings of the European conference on computer vision
  (ECCV)}, 347--363.

\bibitem[{Shetty and Laaksonen(2016)}]{shetty2016frame}
Shetty, R.; and Laaksonen, J. 2016.
\newblock Frame-and segment-level features and candidate pool evaluation for
  video caption generation.
\newblock In \emph{Proceedings of the 24th ACM international conference on
  Multimedia}, 1073--1076.

\bibitem[{Song et~al.(2015)Song, Vallmitjana, Stent, and
  Jaimes}]{song2015tvsum}
Song, Y.; Vallmitjana, J.; Stent, A.; and Jaimes, A. 2015.
\newblock Tvsum: Summarizing web videos using titles.
\newblock In \emph{Proceedings of the IEEE conference on computer vision and
  pattern recognition}, 5179--5187.

\bibitem[{Tang et~al.(2021)Tang, Wang, Liu, Rao, Li, and
  Li}]{tang2021clip4caption}
Tang, M.; Wang, Z.; Liu, Z.; Rao, F.; Li, D.; and Li, X. 2021.
\newblock Clip4caption: Clip for video caption.
\newblock In \emph{Proceedings of the 29th ACM International Conference on
  Multimedia}, 4858--4862.

\bibitem[{Tang et~al.(2002)Tang, Gao, Liu, and Zhang}]{tang2002spatial}
Tang, X.; Gao, X.; Liu, J.; and Zhang, H. 2002.
\newblock A spatial-temporal approach for video caption detection and
  recognition.
\newblock \emph{IEEE transactions on neural networks}, 13(4): 961--971.

\bibitem[{Thomason et~al.(2014)Thomason, Venugopalan, Guadarrama, Saenko, and
  Mooney}]{thomason2014integrating}
Thomason, J.; Venugopalan, S.; Guadarrama, S.; Saenko, K.; and Mooney, R. 2014.
\newblock Integrating language and vision to generate natural language
  descriptions of videos in the wild.
\newblock Technical report, University of Texas at Austin Austin United States.

\bibitem[{Venugopalan et~al.(2016)Venugopalan, Hendricks, Mooney, and
  Saenko}]{venugopalan2016improving}
Venugopalan, S.; Hendricks, L.~A.; Mooney, R.; and Saenko, K. 2016.
\newblock Improving lstm-based video description with linguistic knowledge
  mined from text.
\newblock \emph{arXiv preprint arXiv:1604.01729}.

\bibitem[{Xu et~al.(2019)Xu, Liu, Shu, and Yu}]{xu2019bert}
Xu, H.; Liu, B.; Shu, L.; and Yu, P.~S. 2019.
\newblock BERT post-training for review reading comprehension and aspect-based
  sentiment analysis.
\newblock \emph{arXiv preprint arXiv:1904.02232}.

\bibitem[{Yaliniz and Ikizler-Cinbis(2021)}]{yaliniz2021using}
Yaliniz, G.; and Ikizler-Cinbis, N. 2021.
\newblock Using independently recurrent networks for reinforcement learning
  based unsupervised video summarization.
\newblock \emph{Multimedia Tools and Applications}, 80(12): 17827--17847.

\bibitem[{Yang et~al.(2019)Yang, Dai, Yang, Carbonell, Salakhutdinov, and
  Le}]{yang2019xlnet}
Yang, Z.; Dai, Z.; Yang, Y.; Carbonell, J.; Salakhutdinov, R.~R.; and Le, Q.~V.
  2019.
\newblock Xlnet: Generalized autoregressive pretraining for language
  understanding.
\newblock \emph{Advances in neural information processing systems}, 32.

\bibitem[{Zeng et~al.(2014)Zeng, Liu, Lai, Zhou, and Zhao}]{zeng2014relation}
Zeng, D.; Liu, K.; Lai, S.; Zhou, G.; and Zhao, J. 2014.
\newblock Relation classification via convolutional deep neural network.
\newblock In \emph{Proceedings of COLING 2014, the 25th international
  conference on computational linguistics: technical papers}, 2335--2344.

\bibitem[{Zhang et~al.(2019{\natexlab{a}})Zhang, Guo, Fan, Lan, and
  Cheng}]{zhang2019outline}
Zhang, R.; Guo, J.; Fan, Y.; Lan, Y.; and Cheng, X. 2019{\natexlab{a}}.
\newblock Outline generation: Understanding the inherent content structure of
  documents.
\newblock In \emph{Proceedings of the 42nd International ACM SIGIR Conference
  on Research and Development in Information Retrieval}, 745--754.

\bibitem[{Zhang et~al.(2019{\natexlab{b}})Zhang, Kampffmeyer, Zhao, and
  Tan}]{zhang2019dtr}
Zhang, Y.; Kampffmeyer, M.; Zhao, X.; and Tan, M. 2019{\natexlab{b}}.
\newblock Dtr-gan: Dilated temporal relational adversarial network for video
  summarization.
\newblock In \emph{Proceedings of the ACM Turing Celebration Conference-China},
  1--6.

\bibitem[{Zhang et~al.(2022)Zhang, Li, Bao, Li, Zhang, Li, Huang, and
  Zhang}]{zhang2022mucgec}
Zhang, Y.; Li, Z.; Bao, Z.; Li, J.; Zhang, B.; Li, C.; Huang, F.; and Zhang, M.
  2022.
\newblock MuCGEC: a Multi-Reference Multi-Source Evaluation Dataset for Chinese
  Grammatical Error Correction.
\newblock \emph{arXiv preprint arXiv:2204.10994}.

\bibitem[{Zhang et~al.(2021)Zhang, Qi, Yuan, Shan, Li, Deng, and
  Hu}]{zhang2021open}
Zhang, Z.; Qi, Z.; Yuan, C.; Shan, Y.; Li, B.; Deng, Y.; and Hu, W. 2021.
\newblock Open-book video captioning with retrieve-copy-generate network.
\newblock In \emph{Proceedings of the IEEE/CVF Conference on Computer Vision
  and Pattern Recognition}, 9837--9846.

\bibitem[{Zhou, Qiao, and Xiang(2018)}]{zhou2018deep}
Zhou, K.; Qiao, Y.; and Xiang, T. 2018.
\newblock Deep reinforcement learning for unsupervised video summarization with
  diversity-representativeness reward.
\newblock In \emph{Proceedings of the AAAI Conference on Artificial
  Intelligence}, volume~32.

\bibitem[{Zhou et~al.(2020)Zhou, Palangi, Zhang, Hu, Corso, and
  Gao}]{zhou2020unified}
Zhou, L.; Palangi, H.; Zhang, L.; Hu, H.; Corso, J.; and Gao, J. 2020.
\newblock Unified vision-language pre-training for image captioning and vqa.
\newblock In \emph{Proceedings of the AAAI Conference on Artificial
  Intelligence}, volume~34, 13041--13049.

\bibitem[{Zhu et~al.(2020)Zhu, Lu, Li, and Zhou}]{zhu2020dsnet}
Zhu, W.; Lu, J.; Li, J.; and Zhou, J. 2020.
\newblock Dsnet: A flexible detect-to-summarize network for video
  summarization.
\newblock \emph{IEEE Transactions on Image Processing}, 30: 948--962.

\end{thebibliography}
\appendix
\section{Appendix}
\subsection{Dataset Details}
\begin{figure}[htbp]
    \centering
    \includegraphics[width=\linewidth]{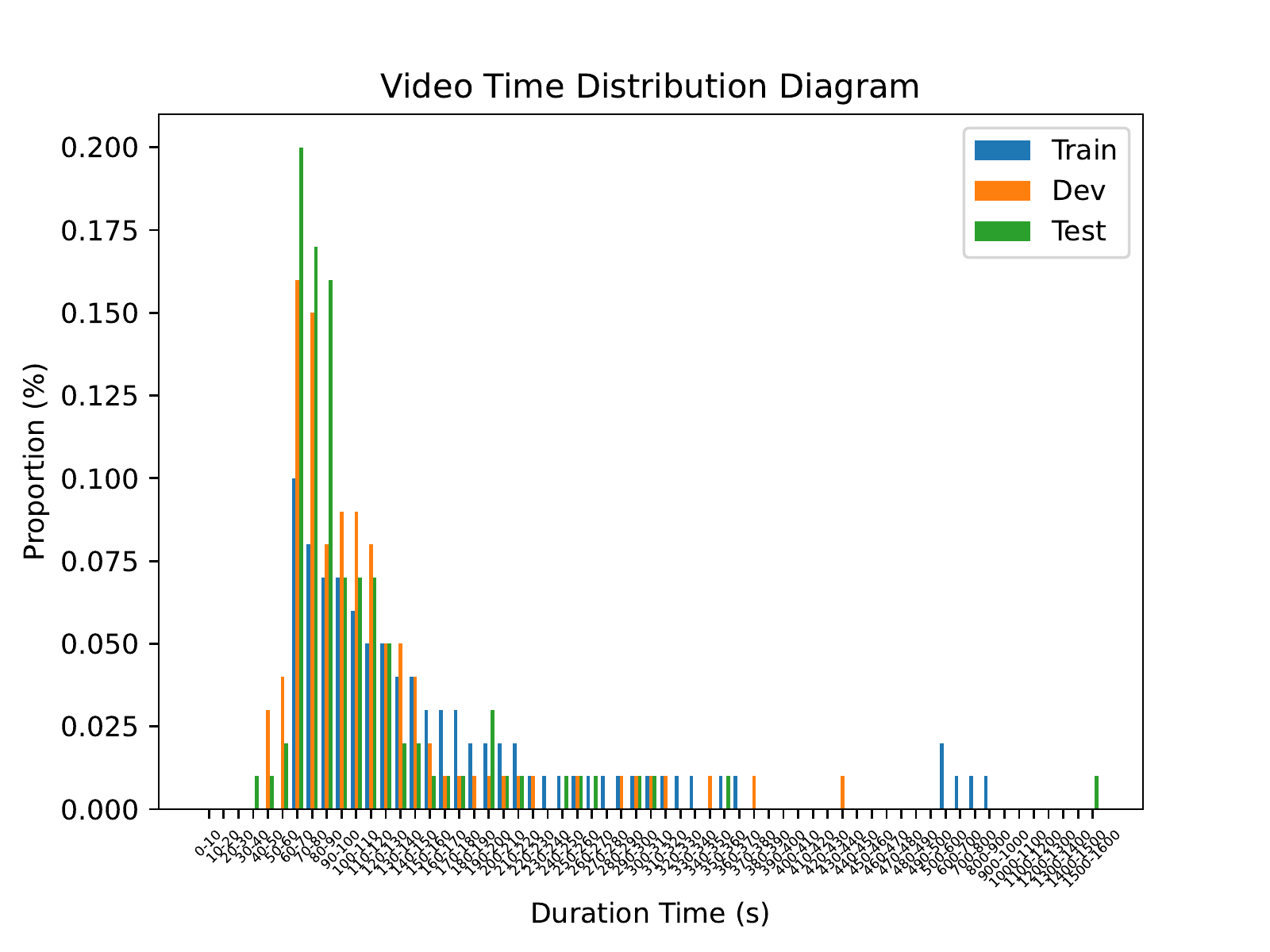}
    \includegraphics[width=\linewidth]{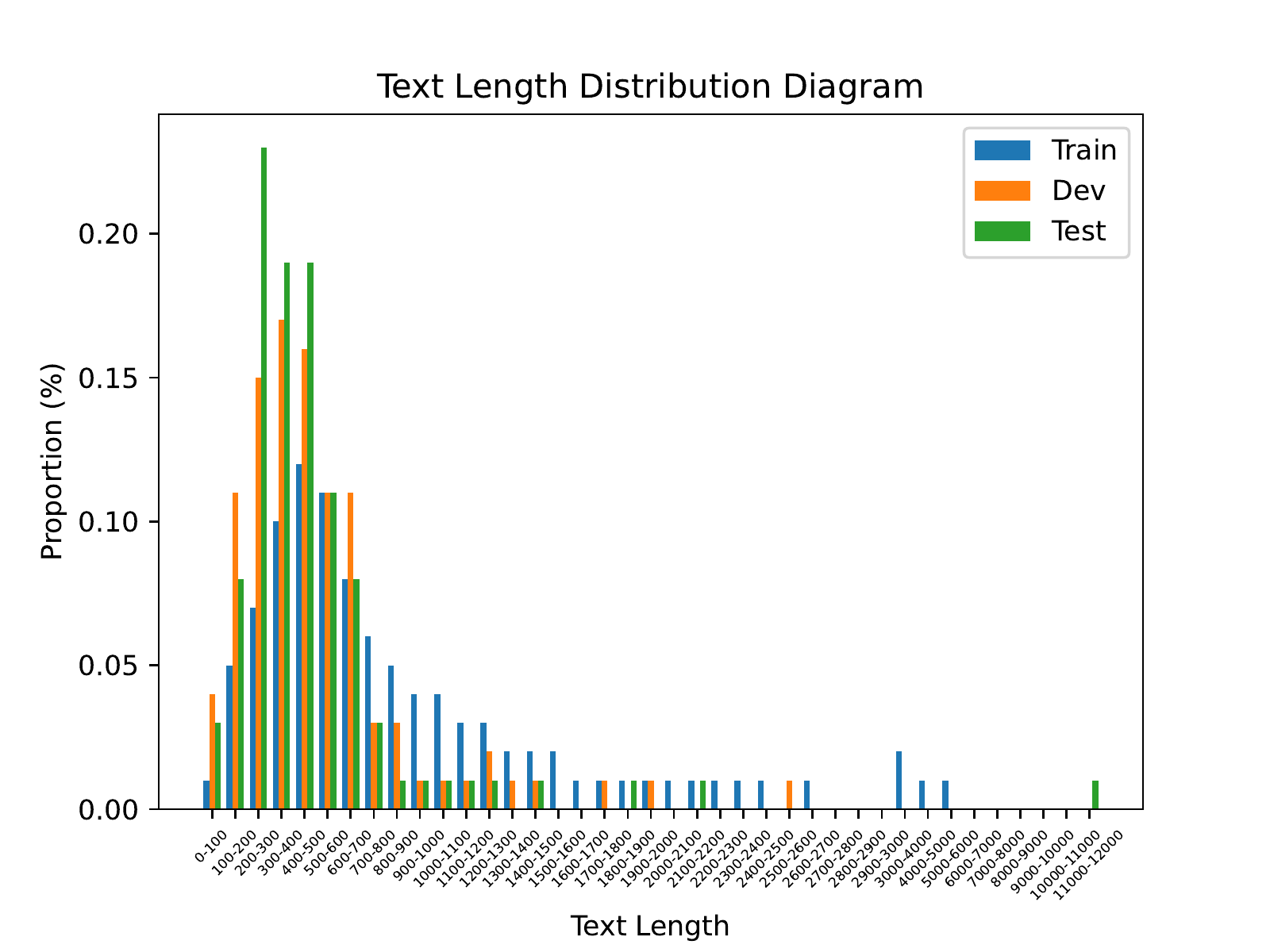}
    \caption{Video time and text length distribution diagram.}
    \label{fig:dataset_text_len}
\end{figure}
Figure \ref{fig:dataset_text_len} shows the video time distribution of our DuVOG dataset, as well as the text length distribution.
Most video lasts between 40 seconds and 150 seconds, while the text length is distributed centrally in the range of 0-800.
It can be seen that our DuVOG dataset is dominated by short-time and medium-time videos.

\subsection{Supplementary Experiment}
We also conduct the sentence-level experiment, namely sentence extraction and rewriting.
Specifically,  similar to the span extraction and rewriting, we extract sentences which contain headlines and rewrite them to formulate the final outlines.
For sentence extraction, we generate the tag for each subtitle sentence.
The tag is only labeled on the comma which replaces the original time point.
Considering the headline would cross different subtitles, we adopt the tags from \{B,I,O\}.
We leverage the BERT as our base model and integrate the CRF layer to predict the result.
For rewriting, we feed the extracted sentences to LaserTagger model, to predict the final outlines.
The settings, evaluation metrics and training hyperparameters are the same as that mentioned before.

\subsubsection{Experiment Result}
Table \ref{tab:ex} shows the comparison between the result of span-level based method \textbf{BC} and sentence-level based method \textbf{Sent-BC}.
It can observed that Sent-BC has a better segmentation performance but is weak in terms of generation.
We believe that for segmentation, the sentences contain more global structural information, while for generation, it is more difficult to retain an important part from a complete sentence than from a span.
Overall, a sentence-level based approach would be weaker.
\begin{table}[htbp]
    \centering
    \small
    \begin{tabular}{|l|ccc|c|c|}
    \hline
          & \multicolumn{3}{c|}{Segmentation} & \multicolumn{1}{c|}{Generation} & \multicolumn{1}{l|}{Overall} \\
    \hline
          & Pre.  & Rec.  & F1   & ROUGE-L$_{F0.5}$ & Score \\
    \hline
    BC    & 79.0  & \textbf{68.9 } & 73.6  & \textbf{84.1 } & \textbf{61.9 } \\
    \hline
    Sent-BC & \textbf{82.9 } & 67.2  & \textbf{74.2 } & 81.8  & 60.7  \\
    \hline
    \end{tabular}%
    \caption{Performance of sentence-level }
    \label{tab:ex}
\end{table}
\end{document}